\documentclass[runningheads]{llncs}
\usepackage{graphicx}
\usepackage{amsmath,amssymb} 
\usepackage{color}
\usepackage[width=122mm,left=12mm,paperwidth=146mm,height=193mm,top=12mm,paperheight=217mm]{geometry}

\begin{document}
\pagestyle{headings}
\mainmatter

\title{Learning to Track at 100 FPS with Deep Regression Networks} 

\titlerunning{Learning to Track}

\authorrunning{Held, Thrun, Savarese}

\author{David Held, Sebastian Thrun, Silvio Savarese}

\institute{Department of Computer Science\\
	Stanford University\\
	\email{ \{davheld,thrun,ssilvio\}@cs.stanford.edu}
}

\maketitle

\begin{abstract}
Machine learning techniques are often used in computer vision due to their ability to leverage large amounts of training data to improve performance.  Unfortunately, most generic object trackers are still trained from scratch online and do not benefit from the large number of videos that are readily available for offline training.  We propose a method for offline training of neural networks that can track novel objects at test-time at 100 fps.  Our tracker is significantly faster than previous methods that use neural networks for tracking, which are typically very slow to run and not practical for real-time applications.  
Our tracker uses a simple feed-forward network with no online training required.
The tracker learns a generic relationship between object motion and appearance and can be used to track novel objects that do not appear in the training set.  We test our network on a standard tracking benchmark to demonstrate our tracker's state-of-the-art performance.  Further, our performance improves as we add more videos to our offline training set.  To the best of our knowledge, our tracker\footnote{Our tracker is available at \url{http://davheld.github.io/GOTURN/GOTURN.html}} is the first neural-network tracker that learns to track generic objects at 100 fps.
\end{abstract}

\section{Introduction}
Given some object of interest marked in one frame of a video, the goal of ``single-target tracking" is to locate this object in subsequent video frames, despite object motion, changes in viewpoint, lighting changes, or other variations.  Single-target tracking is an important component of many systems.  For person-following applications, a robot must track a person as they move through their environment.  For autonomous driving, a robot must track dynamic obstacles in order to estimate where they are moving and predict how they will move in the future.

Generic object trackers (trackers that are not specialized for specific classes of objects) are traditionally trained entirely from scratch online (i.e. during test time) \cite{hare2011struck,babenko2009visual,wang2015understanding,kalal2012tracking}, with no offline training being performed.  Such trackers suffer in performance because they cannot take advantage of the large number of videos that are readily available to improve their performance.  Offline training videos can be used to teach the tracker to handle rotations, changes in viewpoint, lighting changes, and other complex challenges.

In many other areas of computer vision, such as image classification, object detection, segmentation, or activity recognition, machine learning has allowed vision algorithms to train from offline data and learn about the world~\cite{bo2013multipath,krizhevsky2012imagenet,girshick2014rich,leviage,donahue2014long,long2015fully}.  In each of these cases, the performance of the algorithm improves as it iterates through the training set of images.  Such models benefit from the ability of neural networks to learn complex functions from large amounts of data.

\begin{figure}[t]
	\begin{center}
		\includegraphics[width=0.95\linewidth]{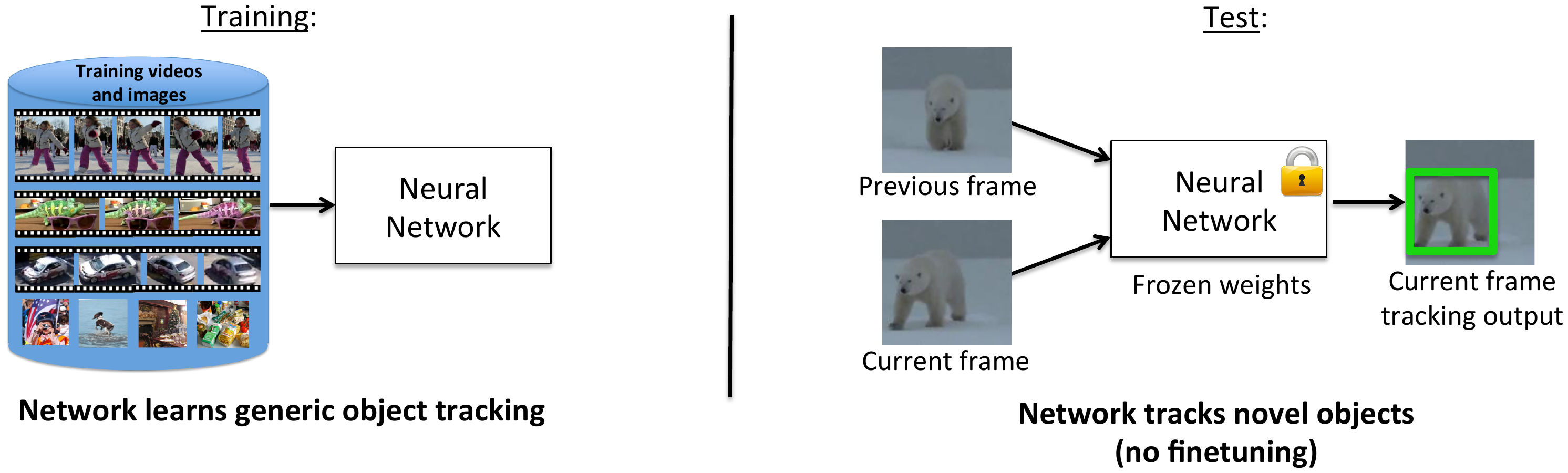}
	\end{center}
	\caption{Using a collection of videos and images with bounding box labels (but no class information), we train a neural network to track generic objects.  At test time, the network is able to track novel objects without any fine-tuning.  By avoiding fine-tuning, our network is able to track at 100 fps}
	\label{fig:pull_track}
\end{figure}

In this work, we show that it is possible to learn to track generic objects in real-time by watching videos offline of objects moving in the world.  To achieve this goal, we introduce \textit{GOTURN}, Generic Object Tracking Using Regression Networks.  We train a neural network for tracking in an entirely offline manner.  At test time, when tracking novel objects, the network weights are frozen, and no online fine-tuning required (as shown in Figure~\ref{fig:pull_track}).  Through the offline training procedure, the tracker learns to track novel objects in a fast, robust, and accurate manner.   

Although some initial work has been done in using neural networks for tracking, these efforts have produced neural-network trackers that are too slow for practical use.  In contrast, our tracker is able to track objects at 100 fps, making it, to the best of our knowledge, the fastest neural-network tracker to-date.  Our real-time speed is due to two factors.  First, most previous neural network trackers are trained online~\cite{li122014deeptrack,li2015deeptrack,wang2015visual,wang2013learning,wang2015transferring,nam2015learning,zhang2015robust,danelljan2015learning,kuen2015self}; however, training neural networks is a slow process, leading to slow tracking.  In contrast, our tracker is trained offline to learn a generic relationship between appearance and motion, so no online training is required.  Second, most trackers take a classification-based approach, classifying many image patches to find the target object~\cite{li122014deeptrack,li2015deeptrack,wang2013learning,nam2015learning,zhang2015robust,kuen2015self,tao2016sint}.  In contrast, our tracker uses a regression-based approach, requiring just a single feed-forward pass through the network to regresses directly to the location of the target object.  The combination of offline training and one-pass regression leads to a significant speed-up compared to previous approaches and allows us to track objects at real-time speeds.  

GOTURN is the first generic object neural-network tracker that is able to run at 100 fps.  We use a standard tracking benchmark to demonstrate that our tracker outperforms state-of-the-art trackers.  Our tracker trains from a set of labeled training videos and images, but we do not require any class-level labeling or information about the types of objects being tracked.  GOTURN establishes a new framework for tracking in which the relationship between appearance and motion is learned offline in a generic manner.  Our code and additional experiments can be found at \url{http://davheld.github.io/GOTURN/GOTURN.html}.

\section{Related Work}
\textbf{Online training for tracking.}
Trackers for generic object tracking are typically trained entirely online, starting from the first frame of a video~\cite{hare2011struck,babenko2009visual,wang2015understanding,kalal2012tracking}.  A typical tracker will sample patches near the target object, which are considered as ``foreground"~\cite{babenko2009visual}.  Some patches farther from the target object are also sampled, and these are considered as ``background."  These patches are then used to train a foreground-background classifier, and this classifier is used to score patches from the next frame to estimate the new location of the target object~\cite{wang2015understanding,kalal2012tracking}.  Unfortunately, since these trackers are trained entirely online, they cannot take advantage of the large amount of videos that are readily available for offline training that can potentially be used to improve their performance.

Some researchers have also attempted to use neural networks for tracking within the traditional online training framework~\cite{li122014deeptrack,li2015deeptrack,wang2015visual,wang2013learning,wang2015transferring,nam2015learning,zhang2015robust,danelljan2015learning,kuen2015self,hong2015tracking}, 
showing state-of-the-art results~\cite{nam2015learning,danelljan2015learning,kristan2015visual}. 
Unfortunately, neural networks are very slow to train, and if online training is required, then the resulting tracker will be very slow at test time.  Such trackers range from 0.8 fps~\cite{li122014deeptrack} to 15 fps~\cite{wang2013learning}, with the top performing neural-network trackers running at 1 fps on a GPU~\cite{nam2015learning,danelljan2015learning,kristan2015visual}.  Hence, these trackers are not usable for most practical applications.  Because our tracker is trained offline in a generic manner, no online training of our tracker is required, enabling us to track at 100 fps.   

\textbf{Model-based trackers.}
A separate class of trackers are the model-based trackers which are designed to track a specific class of objects~\cite{Geiger2013,andriluka2008people,fan2010human}.  For example, if one is only interested in tracking pedestrians, then one can train a pedestrian detector.  During test-time, these detections can be linked together using temporal information.  These trackers are trained offline, but they are limited because they can only track a specific class of objects.  Our tracker is trained offline in a generic fashion and can be used to track novel objects at test time.

\textbf{Other neural network tracking frameworks.}
A related area of research is patch matching~\cite{han2015matchnet,zagoruyko2015learning}, which was recently used for tracking in~\cite{tao2016sint}, running at 4 fps.  In such an approach, many candidate patches are passed through the network, and the patch with the highest matching score is selected as the tracking output.  In contrast, our network only passes two images through the network, and the network regresses directly to the bounding box location of the target object.  By avoiding the need to score many candidate patches, we are able to track objects at 100 fps.

Prior attempts have been made to use neural networks for tracking in various other ways~\cite{jin2013tracking}, including visual attention models~\cite{bazzani2011learning,mnih2014recurrent}. However, these approaches are not competitive with other state-of-the-art trackers when evaluated on difficult tracker datasets. 

\section{Method}

\subsection{Method Overview}
At a high level, we feed frames of a video into a neural network, and the network successively outputs the location of the tracked object in each frame.  We train the tracker entirely offline with video sequences and images.  Through our offline training procedure, our tracker learns a generic relationship between appearance and motion that can be used to track novel objects at test time with no online training required.

\subsection{Input / output format}
\label{sec: Input format}
\begin{figure}[t]
	\begin{center}
		\includegraphics[width=0.6\linewidth]{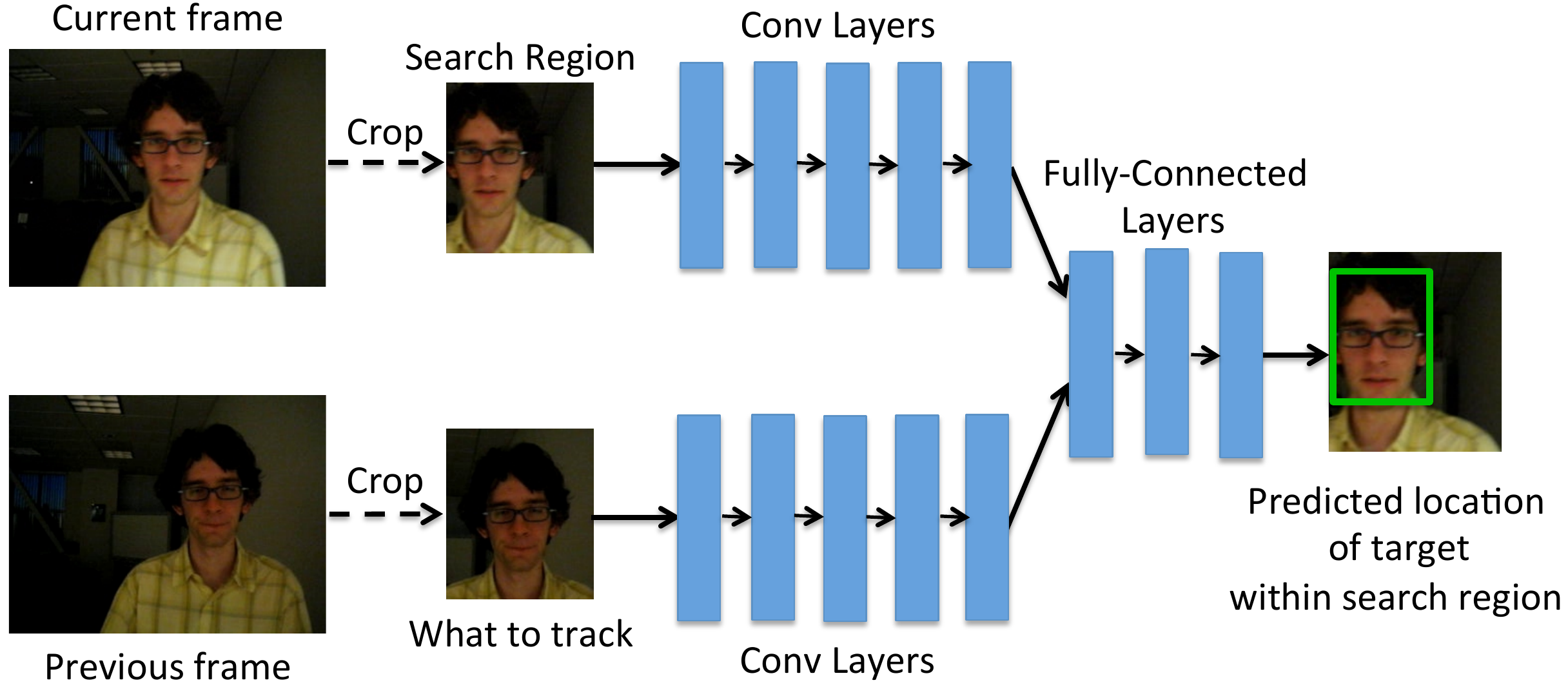}
	\end{center}
	\caption{Our network architecture for tracking.  We input to the network a search region from the current frame and a target from the previous frame.  The network learns to compare these crops to find the target object in the current image}
	\label{fig:network}
\end{figure}

\textbf{What to track.} In case there are multiple objects in the video, the network must receive some information about which object in the video is being tracked.  To achieve this, we input an image of the target object into the network.  We crop and scale the previous frame to be centered on the target object, as shown in Figure~\ref{fig:network}.  This input allows our network to track novel objects that it has not seen before; the network will track whatever object is being input in this crop.
We pad this crop to allow the network to receive some contextual information about the surroundings of the target object.  

In more detail, suppose that in frame $t-1$, our tracker previously predicted that the target was located in a bounding box centered at $c = (c_x, c_y)$ with a width of $w$ and a height of $h$.  At time $t$, we take a crop of frame $t-1$ centered at $(c_x, c_y)$ with a width and height of $k_1 \, w$ and $k_1 \, h$, respectively.  This crop tells the network which object is being tracked.  The value of $k_1$ determines how much context the network will receive about the target object from the previous frame.  

\textbf{Where to look.}
To find the target object in the current frame, the tracker should know where the object was previously located.  Since objects tend to move smoothly through space, the previous location of the object will provide a good guess of where the network should expect to currently find the object.  We achieve this by choosing a search region in our current frame based on the object's previous location.  We crop the current frame using the search region and input this crop into our network, as shown in Figure~\ref{fig:network}.  The goal of the network is then to regress to the location of the target object within the search region.  

In more detail, the crop of the current frame $t$ is centered at $c' = (c'_x, c'_y)$, where $c'$ is the expected mean location of the target object.   We  set $c' = c$, which is equivalent to a constant position motion model, although more sophisticated motion models can be used as well.  The crop of the current frame has a width and height of $k_2 \, w$ and $k_2 \, h$, respectively, where $w$ and $h$ are the width and height of the predicted bounding box in the previous frame, and $k_2$ defines our search radius for the target object.  In practice, we use $k_1 = k_2 = 2$.  

As long as the target object does not become occluded and is not moving too quickly, the target will be located within this region.  For fast-moving objects, the size of the search region could be increased, at a cost of increasing the complexity of the network.  Alternatively, to handle long-term occlusions or large movements, our tracker can be combined with another approach such as an online-trained object detector, as in the TLD framework~\cite{kalal2012tracking}, or a  visual attention model~\cite{bazzani2011learning,mnih2014recurrent,ba2014multiple}; we leave this for future work.

\textbf{Network output.}
The network outputs the coordinates of the object in the current frame, relative to the search region.  The network's output consists of the coordinates of the top left and bottom right corners of the bounding box.  

\subsection{Network architecture}
\label{sec:Two-stream model}
For single-target tracking, we define a novel image-comparison tracking architecture, shown in Figure~\ref{fig:network} (note that related ``two-frame" architectures have also been used for other tasks~\cite{karpathy2014large,dosovitskiy2015flownet}).  In this model, we input the target object as well as the search region each into a sequence of convolutional layers.  The output of these convolutional layers is a set of features that capture a high-level representation of the image.  

The outputs of these convolutional layers are then fed through a number of fully connected layers.  
The role of the fully connected layers is to compare the features from the target object to the features in the current frame to find where the target object has moved.  Between these frames, the object may have undergone a translation, rotation, lighting change, occlusion, or deformation.  The function learned by the fully connected layers is thus a complex feature comparison which is learned through many examples to be robust to these various factors while outputting the relative motion of the tracked object.  

In more detail, the convolutional layers in our model are taken from the first five convolutional layers of the CaffeNet architecture~\cite{jia2014caffe,krizhevsky2012imagenet}. 
We concatenate the output of these convolutional layers (i.e. the pool5 features) into a single vector.  This vector is input to 3 fully connected layers, each with 4096 nodes.  Finally, we connect the last fully connected layer to an output layer that contains 4 nodes which represent the output bounding box.  We scale the output by a factor of 10, chosen using our validation set (as with all of our hyperparameters).  Network hyperparameters are taken from the defaults for CaffeNet, and between each fully-connected layer we use dropout and ReLU non-linearities as in CaffeNet.  Our neural network is implemented using Caffe~\cite{jia2014caffe}.  

\subsection{Tracking}
During test time, we initialize the tracker with a ground-truth bounding box from the first frame, as is standard practice for single-target tracking.  At each subsequent frame t, we input crops from frame $t-1$ and frame $t$ into the network (as described in Section~\ref{sec: Input format}) to predict where the object is located in frame $t$.  We continue to re-crop and feed pairs of frames into our network for the remainder of the video, and our network will track the movement of the target object throughout the entire video sequence.

\section{Training}
\label{sec:Training}
We train our network with a combination of videos and still images.  The training procedure is described below.  In both cases, we train the network with an L1 loss between the predicted bounding box and the ground-truth bounding box.  

\subsection{Training from Videos and Images}
\label{sec:Training Data}
\begin{figure}[t]
	\begin{center}
		\includegraphics[width=0.7\linewidth]{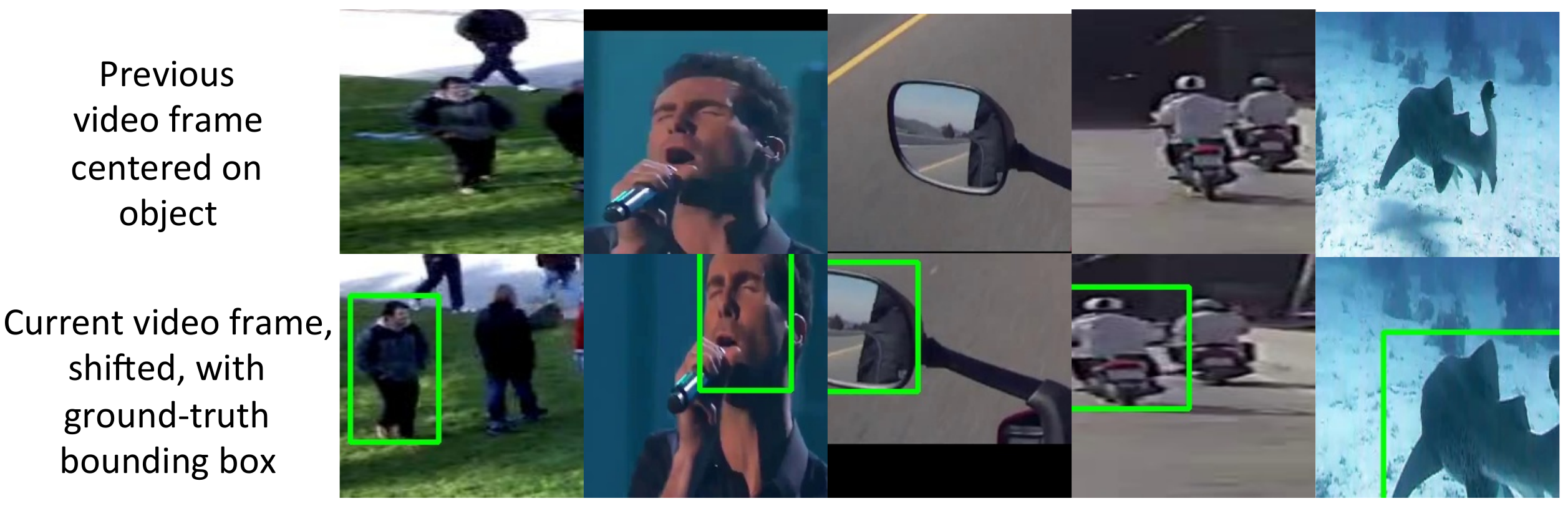}
	\end{center}
	\caption{Examples of training videos.  The goal of the network is to predict the location of the target object shown in the center of the video frame in the top row, after being shifted as in the bottom row.  The ground-truth bounding box is marked in green}
	\label{fig:video_crops}
\end{figure}

Our training set consists of a collection of videos in which a subset of frames in each video are labeled with the location of some object.  For each successive pair of frames in the training set, we crop the frames as described in Section~\ref{sec: Input format}.  During training time, we feed this pair of frames into the network and attempt to predict how the object has moved from the first frame to the second frame (shown in Figure~\ref{fig:video_crops}).  We also augment these training examples using our motion model, as described in Section~\ref{sec: Random cropping}.

\begin{figure}[t]
	\begin{center}
		\includegraphics[width=0.7\linewidth]{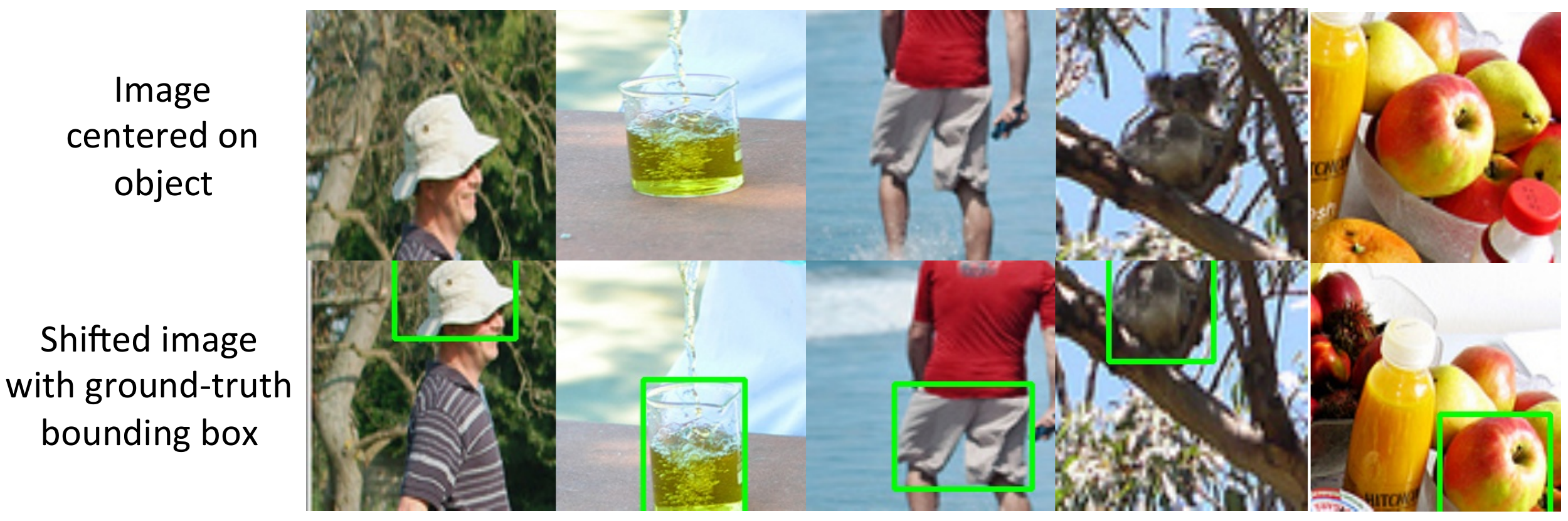}
	\end{center}
	\caption{Examples of training images.  The goal of the network is to predict the location of the target object shown in the center of the image crop in the top row, after being shifted as in the bottom row.  The ground-truth bounding box is marked in green}
	\label{fig:image_crops}
\end{figure}

Our training procedure can also take advantage of a set of still images that are each labeled with the location of an object.  This training set of images teaches our network to track a more diverse set of objects and prevents overfitting to the objects in our training videos.
To train our tracker from an image, we take random crops of the image according to our motion model (see Section~\ref{sec: Random cropping}).   Between these two crops, the target object has undergone an apparent translation and scale change, as shown in  Figure~\ref{fig:image_crops}.  We treat these crops as if they were taken from different frames of a video.   Although the ``motions" in these crops are less varied than the types of motions found in our training videos, these images are still useful to train our network to track a variety of different objects.

\subsection{Learning Motion Smoothness}
\label{sec: Random cropping}
Objects in the real-world tend to move smoothly through space.  Given an ambiguous image in which the location of the target object is uncertain, a tracker should predict that the target object is located near to the location where it was previously observed.  This is especially important in videos that contain multiple nearly-identical objects, such as multiple fruit of the same type.  Thus we wish to teach our network that, all else being equal, small motions are preferred to large motions. 

To concretize the idea of motion smoothness, we model the center of the bounding box in the current frame $ (c'_x, c'_y)$ relative to the center of the bounding box in the previous frame $(c_x, c_y)$ as  
\begin{align}
	\label{eq: delta_pos1}
	c'_x &= c_x + w \cdot \Delta x \\ 
	\label{eq: delta_pos2}
	c'_y &= c_y + h \cdot \Delta y 
\end{align}
where $w$ and $h$ are the width and height, respectively, of the bounding box of the previous frame.  The terms $\Delta x$ and $\Delta y$ are random variables that capture the change in position of the bounding box relative to its size.
In our training set, we find that objects change their position such that $\Delta x$ and $\Delta y$ can each be modeled with a Laplace distribution with a mean of 0 (see Appendix for details).  Such a distribution places a higher probability on smaller motions than larger motions.

Similarly, we model size changes by 
\begin{align}
	\label{eq: delta_pos3}
	w' &= w \cdot \gamma_w \\
	\label{eq: delta_pos4}
	h' &= h \cdot \gamma_h
\end{align}
where $w'$ and $h'$ are the current width and height of the bounding box and 
$w$ and $h$ are the previous width and height of the bounding box.  The terms $\gamma_w$ and $\gamma_h$ are random variables that capture the size change of the bounding box.
We find in our training set that $\gamma_w$ and $\gamma_h$ are modeled by a Laplace distribution with a mean of 1.  
Such a distribution gives a higher probability on keeping the bounding box size near the same as the size from the previous frame.

To teach our network to prefer small motions to large motions, we augment our training set with random crops drawn from the Laplace distributions described above (see Figures~\ref{fig:video_crops} and~\ref{fig:image_crops} for examples).  Because these training examples are sampled from a Laplace distribution, small motions will be sampled more than large motions, and thus our network will learn to prefer small motions to large motions, all else being equal.  We will show that this Laplace cropping procedure improves the performance of our tracker compared to the standard uniform cropping procedure used in classification tasks~\cite{krizhevsky2012imagenet}.

The scale parameters for the Laplace distributions are chosen via cross-validation to be $b_x = 1/5$ (for the motion of the bounding box center) and $b_s = 1/15$ (for the change in bounding box size).
We constrain the random crop such that it must contain at least half of the target object in each dimension.  
We also limit the size changes such that $\gamma_w, \gamma_h \in (0.6, 1.4)$, to avoid overly stretching or shrinking the bounding box in a way that would be difficult for the network to learn.  

\subsection{Training procedure}
To train our network, each training example is alternately taken from a video or from an image.  When we use a video training example, we randomly choose a video, and we randomly choose a pair of successive frames in this video.  We then crop the video according to the procedure described in Section~\ref{sec: Input format}.  We additionally take $k_3$ random crops of the current frame, as described in Section~\ref{sec: Random cropping}, to augment the dataset with $k_3$ additional examples.  Next, we randomly sample an image, and we repeat the procedure described above, where the random cropping creates artificial ``motions" (see Sections~\ref{sec:Training Data} and~\ref{sec: Random cropping}).  Each time a video or image gets sampled, new random crops are produced on-the-fly, to create additional diversity in our training procedure.  In our experiments, we use $k_3 = 10$, and we use a batch size of 50.  

The convolutional layers in our network are pre-trained on ImageNet~\cite{russakovsky2014imagenet,deng2009imagenet}.  Because of our limited training set size, we do not fine-tune these layers to prevent overfitting.  
We train this network with a learning rate of 1e-5, and other hyperparameters are taken from the defaults for CaffeNet~\cite{jia2014caffe}.  

\section{Experimental Setup}

\subsection{Training set}
As described in Section~\ref{sec:Training}, we train our network using a combination of videos and still images.  Our training videos come from ALOV300++~\cite{smeulders2014visual}, a collection of 314 video sequences.  We remove 7 of these videos that overlap with our test set (see Appendix for details), leaving us with 307 videos to be used for training.  In this dataset, approximately every 5th frame of each video has been labeled with the location of some object being tracked.  These videos are generally short, ranging from a few seconds to a few minutes in length.  We split these videos into 251 for training and 56 for validation / hyper-parameter tuning.  The training set consists of a total of 13,082 images of 251 different objects, or an average of 52 frames per object.  The validation set consists of 2,795 images of 56 different objects.  After choosing our hyperparameters, we retrain our model using our entire training set (training + validation).  After removing the 7 overlapping videos, there is no overlap between the videos in the training and test sets.

Our training procedure also leveraged a set of still images that were used for training, as described in Section~\ref{sec:Training Data}.  These images were taken from the training set of the ImageNet Detection Challenge~\cite{russakovsky2014imagenet}, in which 478,807 objects were labeled with bounding boxes.  We randomly crop these images during training time, as described in Section~\ref{sec: Random cropping}, to create an apparent translation or scale change between two random crops.  The random cropping procedure is only useful if the labeled object does not fill the entire image; thus, we filter those images for which the bounding box fills at least 66\% of the size of the image in either dimension (chosen using our validation set).  This leaves us with a total of 239,283 annotations from 134,821 images.  These images help prevent overfitting by teaching our network to track objects that do not appear in the training videos.

\subsection{Test set}
\label{sec: Test set}
Our test set consists of the 25 videos from the VOT 2014 Tracking Challenge~\cite{kristan2014visual}. We could not test our method on the VOT 2015 challenge~\cite{kristan2015visual} because there would be too much overlap between the test set and our training set.  However, we expect the general trends of our method to still hold.

The VOT 2014 Tracking Challenge~\cite{kristan2014visual} is a standard tracking benchmark that allows us to compare our tracker to a wide variety of state-of-the-art trackers.  The trackers are evaluated using two standard tracking metrics: accuracy ($A$) and robustness ($R$)~\cite{kristan2014visual,cehovin2014my}, which range from 0 to 1.  We also compute accuracy errors $(1-A)$, robustness errors $(1-R)$, and overall errors $1 - (A+R)/2$.  

Each frame of the video is annotated with a number of attributes: occlusion, illumination change,  motion change, size change, and camera motion.  The trackers are also ranked in accuracy and robustness separately for each attribute, and the rankings are then averaged across attributes to get a final average accuracy and robustness ranking for each tracker.  The accuracy and robustness rankings are averaged to get an overall average ranking.  

\section{Results}
\subsection{Overall performance}

\begin{figure}[htb]
	\begin{center}
		\includegraphics[width=0.75\linewidth]{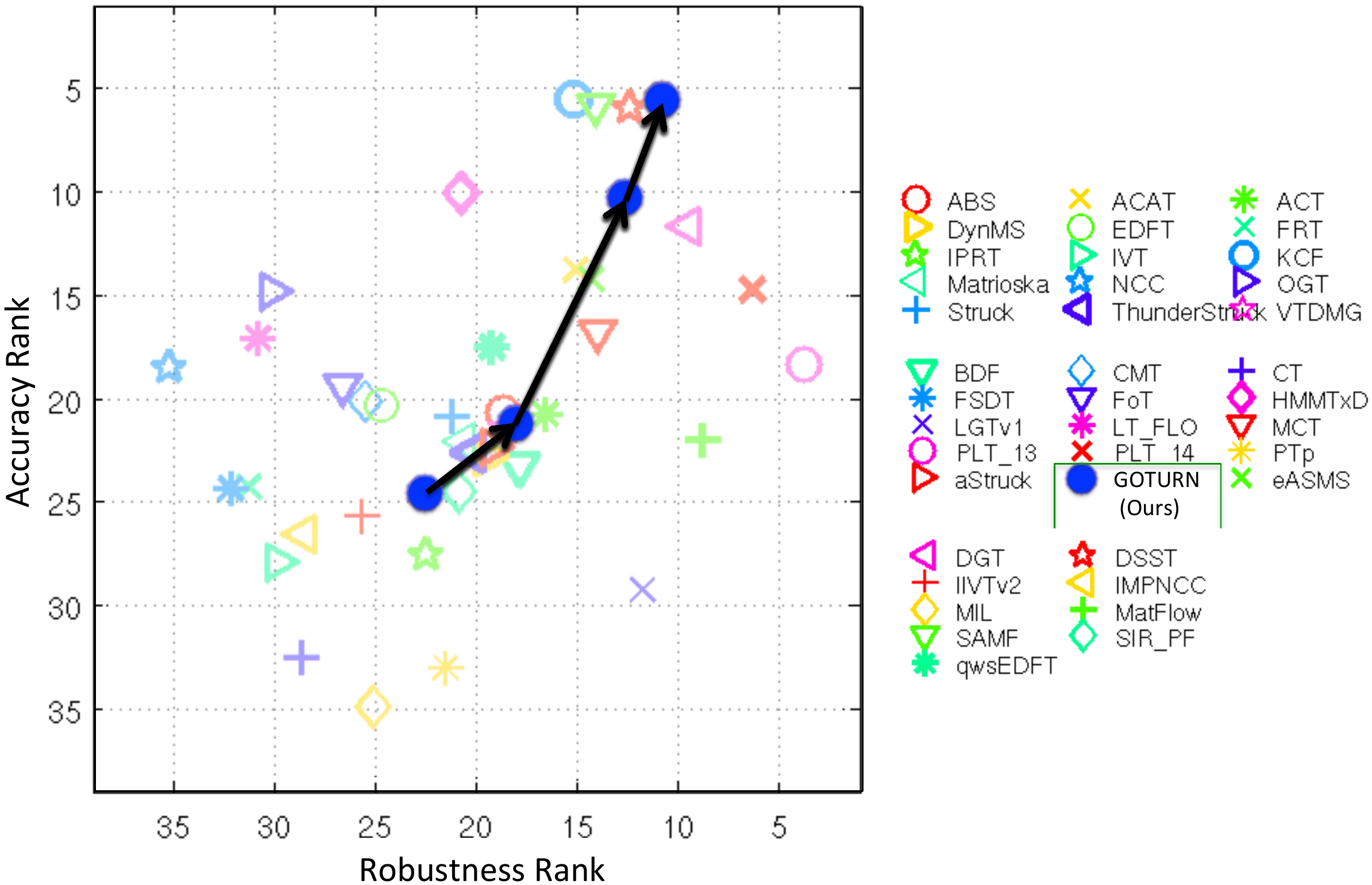}
	\end{center}
	\caption{Tracking results from the VOT 2014 tracking challenge.  Our tracker's performance is indicated with a blue circle, outperforming all previous methods on the overall rank (average of accuracy and robustness ranks).  The points shown along the black line represent training from 14, 37, 157, and 307 videos, with the same number of training images used in each case}
	\label{fig:results}
\end{figure}

The performance of our tracker is shown in Figure~\ref{fig:results}, which demonstrates that our tracker has good robustness and performs near the top in accuracy.  Further, our overall ranking (computed as the average of accuracy and robustness) outperforms all previous trackers on this benchmark.  
We have thus demonstrated the value of offline training for improving tracking performance.  Moreover, these results were obtained after training on only 307 short videos.  Figure~\ref{fig:results} as well as analysis in the appendix suggests that further gains could be achieved if the training set size were increased by labeling more videos.  Qualitative results, as well as failure cases, can be found on the project page: \url{http://davheld.github.io/}; currently, the tracker can fail due to occlusions or overfitting to objects in the training set.

On an Nvidia GeForce GTX Titan X GPU with cuDNN acceleration, our tracker runs at 6.05 ms per frame (not including the 1 ms to load each image in OpenCV), or 165 fps.  On a GTX 680 GPU, our tracker runs at an average of 9.98 ms per frame, or 100 fps.  If only a CPU is available, the tracker runs at 2.7 fps.  Because our tracker is able to perform all of its training offline, during test time the tracker requires only a single feed-forward pass through the network, and thus the tracker is able to run at real-time speeds.  

\begin{figure}[h!]
	\begin{center}
		\includegraphics[width=0.4\linewidth]{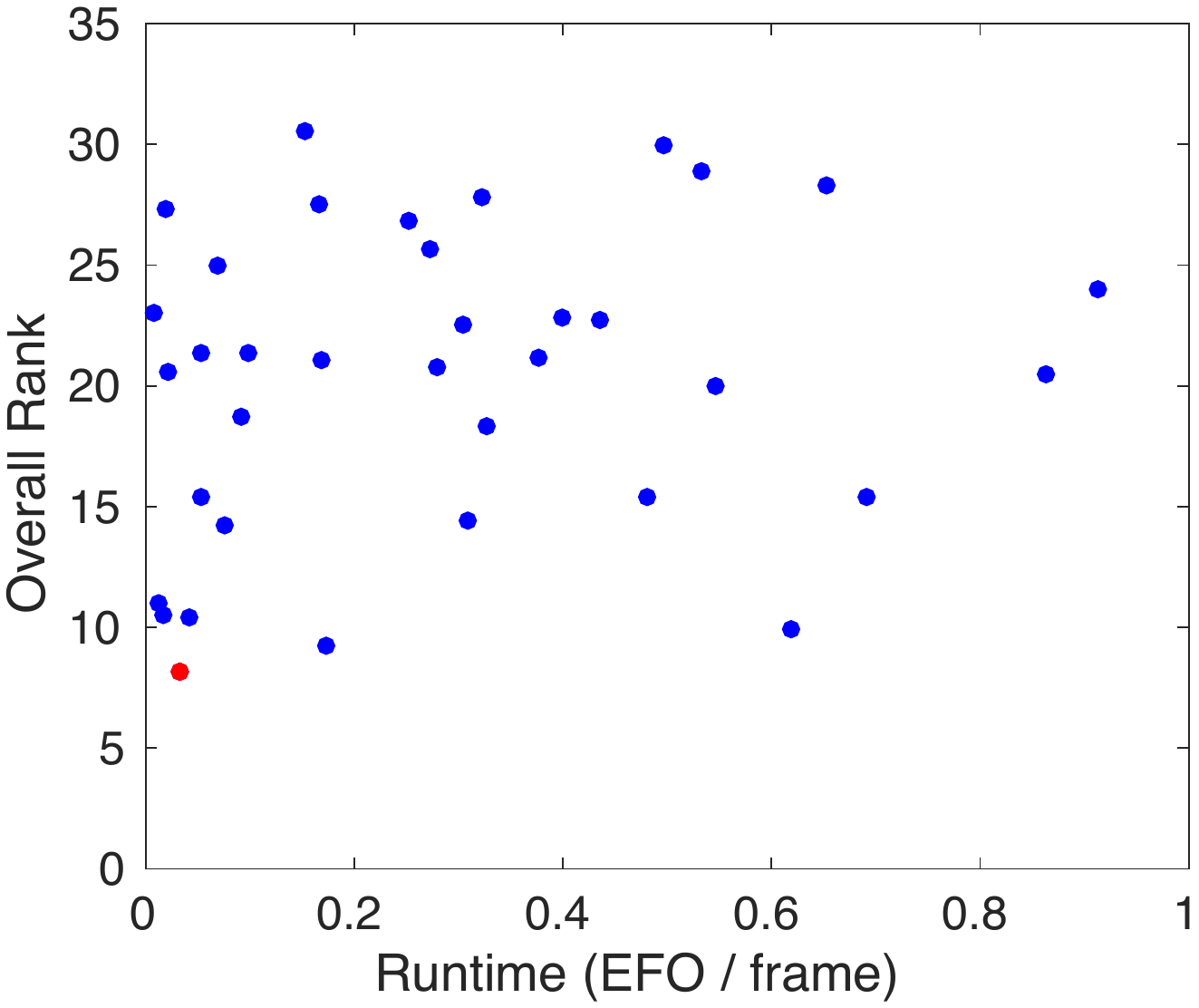}
	\end{center}
	\caption{Rank vs runtime of our tracker (red) compared to the 38 baseline methods from the VOT 2014 Tracking Challenge (blue).  Each blue dot represents the performance of a separate baseline method (best viewed in color).  Accuracy and robustness metrics are shown in the appendix}
	\label{fig:speeds}
\end{figure}

We compare the speed and rank of our tracker compared to the 38 other trackers submitted to the VOT 2014 Tracking Challenge~\cite{kristan2014visual} in Figure~\ref{fig:speeds}, using the overall rank score described in Section~\ref{sec: Test set}.  We show the runtime of the tracker in EFO units (Equivalent Filter Operations), which normalizes for the type of hardware that the tracker was tested on~\cite{kristan2014visual}.  Figure~\ref{fig:speeds} demonstrates that ours was one of the fastest trackers compared to the 38 other baselines, while outperforming all other methods in the overall rank (computed as the average of the accuracy and robustness ranks).  Note that some of these other trackers, such as ThunderStruck~\cite{kristan2014visual}, also use a GPU.  For a more detailed analysis of speed as a function of accuracy and robustness, see the appendix.

\label{sec: Overall performance}
Our tracker is able to track objects in real-time due to two aspects of our model: First, we 
learn a generic tracking model offline, so no online training is required.  Online training of neural networks tends to be very slow, preventing real-time performance.  Online-trained neural network trackers range from 0.8 fps~\cite{li122014deeptrack} to 15 fps~\cite{wang2013learning},  with the top performing trackers running at 1 fps on a GPU~\cite{nam2015learning,danelljan2015learning,kristan2015visual}.
Second, 
most trackers evaluate a finite number of samples and choose the highest scoring one as the tracking output~\cite{li122014deeptrack,li2015deeptrack,wang2013learning,nam2015learning,zhang2015robust,kuen2015self,tao2016sint}.  With a sampling approach, the accuracy is limited by the number of samples, but increasing the number of samples also increases the computational complexity.  On the other hand, our tracker regresses directly to the output bounding box, so GOTURN achieves accurate tracking with no extra computational cost, enabling it to track objects at 100 fps.  

\subsection{How does it work?}
\label{sec: Memory}
How does our neural-network tracker work?  There are two hypotheses that one might propose:
\begin{enumerate}
	\item The network compares the previous frame to the current frame to find the target object in the current frame.
	\item The network acts as a local generic ``object detector" and simply locates the nearest ``object."
\end{enumerate}
We differentiate between these hypotheses by comparing the performance of our network (shown in Figure~\ref{fig:network}) to the performance of a network which does not receive the previous frame as input (i.e. the network only receives the current frame as input).  For this experiment, we train each of these networks separately.  If the network does not receive the previous frame as input, then the tracker can only act as a local generic object detector (hypothesis 2).

\begin{figure}[h!]
	\begin{center}
		\includegraphics[width=0.45\linewidth]{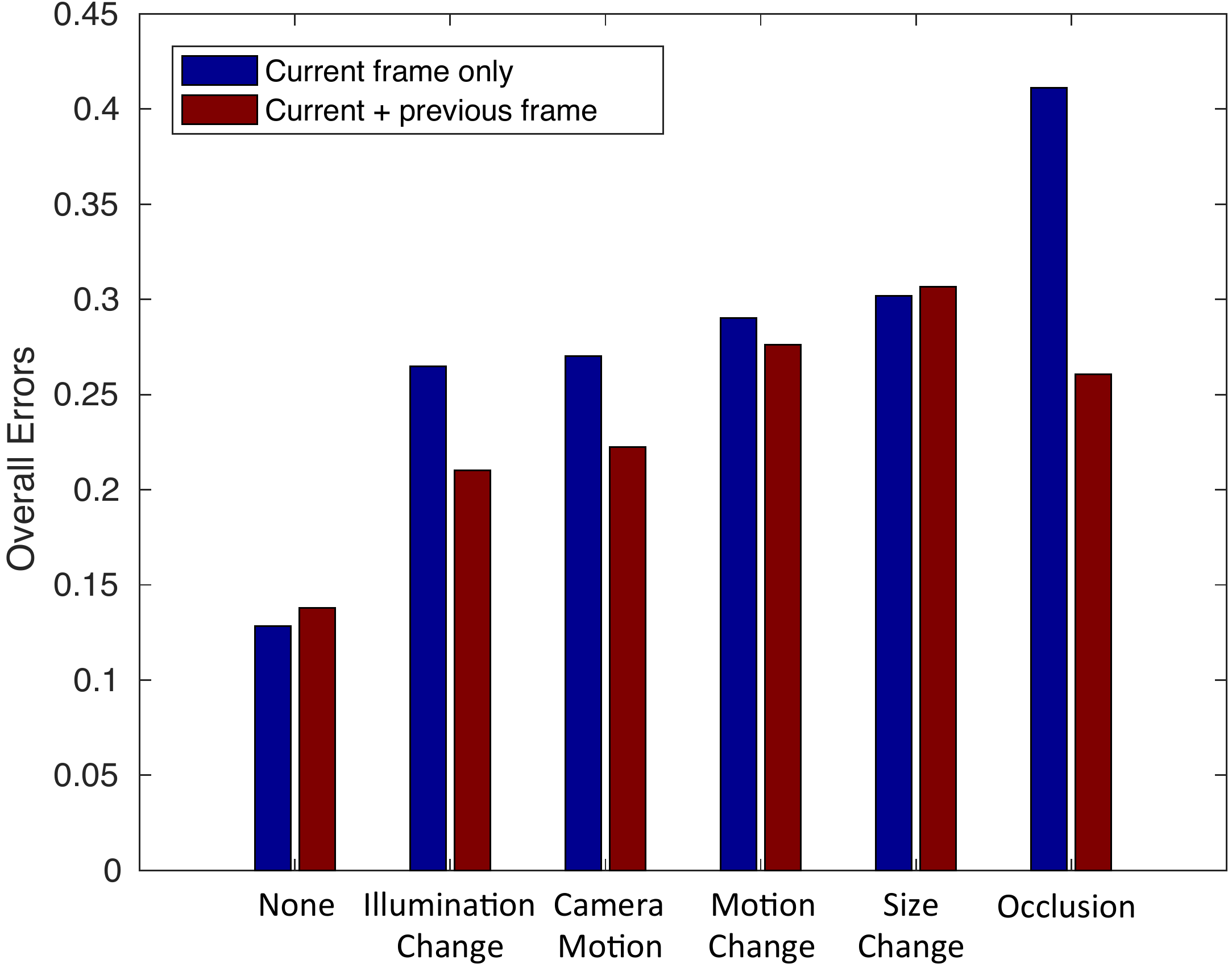}
	\end{center}
	\caption{Overall tracking errors for our network which receives as input both the current and previous frame, compared to a network which receives as input only the current frame (lower is better).  This comparison allows us to disambiguate between two hypotheses that can explain how our neural-network tracker works (see Section~\ref{sec: Memory}).  Accuracy and robustness metrics are shown in the appendix}
	\label{fig:Memory}
\end{figure}

Figure~\ref{fig:Memory} shows the degree to which each of the hypotheses holds true for different tracking conditions.  For example, when there is an occlusion or a large camera motion, the tracker benefits greatly from using the previous frame, which enables the tracker to ``remember" which object is being tracked.  Figure~\ref{fig:Memory} shows that the tracker performs much worse in these cases when the previous frame is not included.  In such cases, hypothesis 1 plays a large role, i.e. the tracker is comparing the previous frame to the current frame to find the target object.  

On the other hand, when there is a size change or no variation, the tracker performs slightly worse when using the previous frame (or approximately the same).  Under a large size change, the corresponding appearance change is too drastic for our network to perform an accurate comparison between the previous frame and the current frame.  Thus the tracker is acting as a local generic object detector in such a case and hypothesis 2 is dominant.  Each hypothesis holds true in varying degrees for different tracking conditions, as shown in Figure~\ref{fig:Memory}.

\subsection{Generality vs Specificity}
\begin{figure}[htb]
	\begin{center}
		\includegraphics[width=0.37\linewidth]{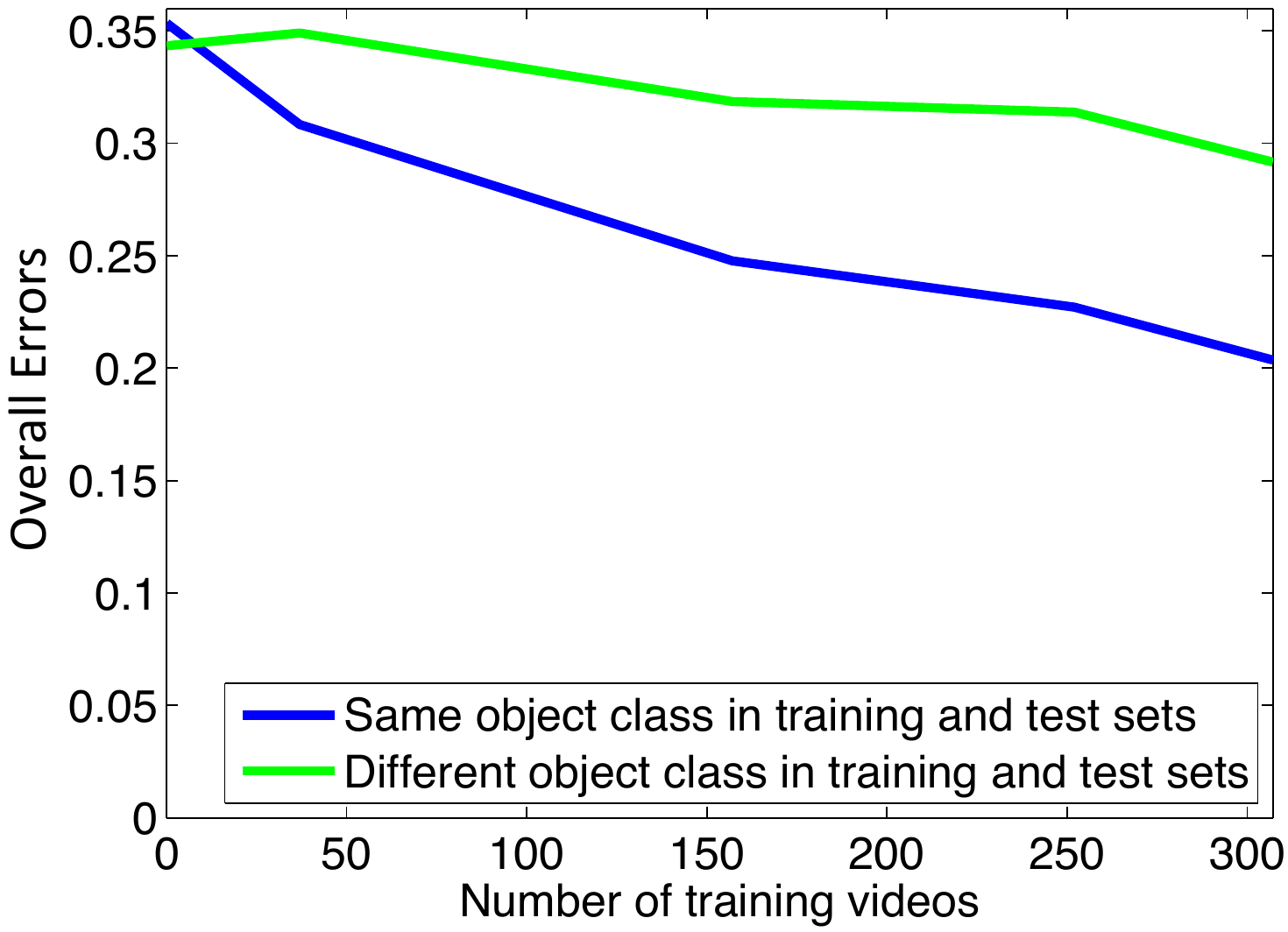}
	\end{center}
	\caption{Overall tracking errors for different types of objects in our test set as a function of the number of videos in our training set (lower is better).  Class labels are not used by our tracker; these labels were obtained only for the purpose of this analysis.  Accuracy and robustness metrics are shown in the appendix}
	\label{fig:Class}
\end{figure}

How well can our tracker generalize to novel objects not found in our training set?  For this analysis, we separate our test set into objects for which at least 25 videos of the same class appear in our training set and objects for which fewer than 25 videos of that class appear in our training set.   Figure~\ref{fig:Class} shows that, even for test objects that do not have any (or very few) similar objects in our training set, our tracker performs well.  The performance continues to improve even as videos of unrelated objects are added to our training set, since our tracker is able to learn a generic relationship between an object's appearance change and its motion that can generalize to novel objects.  

Additionally, our tracker can also be specialized to track certain objects particularly well.  Figure~\ref{fig:Class} shows that, for test objects for which at least 25 videos of the same class appear in the training set, we obtain a large improvement as more training videos of those types of objects are added.  This allows the user to specialize the tracker for particular applications.  For example, if the tracker is being used for autonomous driving, then the user can add more objects of people, bikes, and cars into the training set, and the tracker will learn to track those objects particularly well.  At the same time, Figure~\ref{fig:Class} also demonstrates that our tracker can track novel objects that do not appear in our training set, which is important when tracking objects in uncontrolled environments.

\subsection{Ablative Analysis}
In Table~\ref{tab:comparison}, we show which components of our system contribute the most to our performance.
We train our network with random cropping from a Laplace distribution to teach our tracker to prefer small motions to large motions (e.g. motion smoothness), as explained in Section~\ref{sec: Random cropping}.  Table~\ref{tab:comparison} shows the benefit of this approach compared to the baseline of uniformly sampling random crops (``No motion smoothness"), as is typically done for classification~\cite{krizhevsky2012imagenet}.  As shown, we reduce errors by 20\% by drawing our random crops from a Laplace distribution.  

Table~\ref{tab:comparison} also shows the benefit of using an L1 loss compared to an L2 loss.  Using an L1 loss significantly reduces the overall tracking errors from 0.43 to 0.24.  Because the L2 penalty is relatively flat near 0, the network does not sufficiently penalize outputs that are close but not correct, and the network would often output a bounding box that was slightly too large or too small.  When applied to a sequence of frames, the bounding box would grow or shrink without bound until the predicted bounding box was just a single point or the entire image.  In contrast, an L1 loss penalizes more harshly answers that are only slightly incorrect, which keeps the bounding box size closer to the correct size and prevents the bounding box from shrinking or growing without bound.

\begin{table}[h]
	\begin{center}
		\caption{Comparing our full GOTURN tracking method to various modified versions of our method to analyze the effect of different components of the system}			
		\label{tab:comparison}
		\begin{tabular}{| p{0.3\textwidth} | c | c| c|  }
			\hline
			GOTURN Variant & Overall errors & Accuracy errors & Robustness errors \\
			\hline
			L2 loss & 0.43 & 0.69 & 0.17\\			
			No motion smoothness & 0.30  & 0.48 & 0.13 \\    
			Image training only & 0.35  & 0.54 & 0.16 \\ 
			Video training only & 0.29 & 0.44 & 0.13 \\ 
			\hline
			\textbf{Full method (Ours)} & \textbf{0.24} & \textbf{0.39} & \textbf{0.10} \\
			\hline
		\end{tabular}
	\end{center}
\end{table}

We train our tracker using a combination of images and videos.  Table~\ref{tab:comparison} shows that, given the choice between images and videos, training on only videos gives a much bigger improvement to our tracker performance.  At the same time, training on both videos and images gives the maximum performance for our tracker.  Training on a small number of labeled videos has taught our tracker to be invariant to background motion, out-of-plane rotations, deformations, lighting changes, and minor occlusions.  Training from a large number of labeled images has taught our network how to track a wide variety of different types of objects.  By training on both videos and images, our tracker learns to track a variety of object types under different conditions, achieving maximum performance.

\section{Conclusions}
We have demonstrated that we can train a generic object tracker offline such that its performance improves by watching more training videos.  During test time, we run the network in a purely feed-forward manner with no online fine-tuning required, allowing the tracker to run at 100 fps.    Our tracker learns offline a generic relationship between an object's appearance and its motion, allowing our network to track novel objects at real-time speeds. 

\noindent \textbf{Acknowledgments.}
We acknowledge the support of Toyota grant 1186781-31-UDARO and ONR grant 1165419-10-TDAUZ.

\bibliographystyle{splncs03}
\bibliography{egbib}

\clearpage
\appendix

\section{Offline training}
\label{sec: Results - Amount of training data}
\begin{figure}[h]
	\begin{center}
		\includegraphics[width=0.95\linewidth]{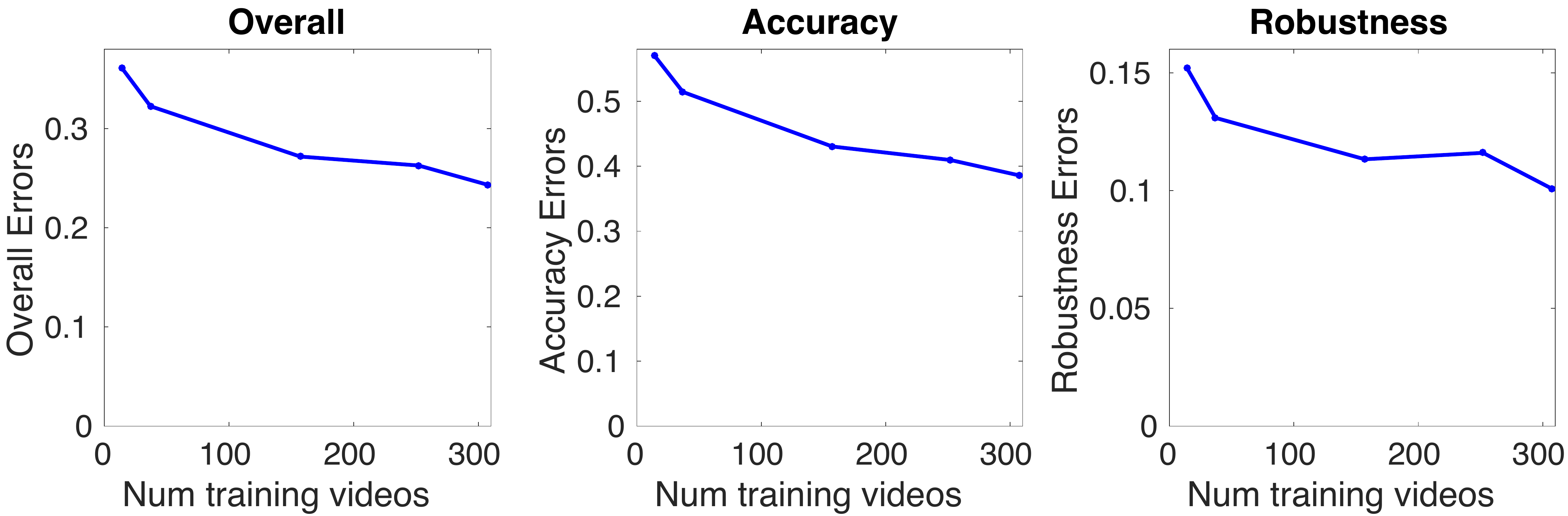}
	\end{center}
	\caption{Tracking performance as a function of the number of training videos (lower is better).  This analysis indicates that large gains are possible by labeling more training videos.}
	\label{fig:num_videos}
\end{figure}

Our tracker is able to improve its performance as it trains on more offline data.  By observing more videos, GOTURN learns how the appearance of objects change as they move.  We further analyze the effect of the amount of training data on our tracker's performance in Figure~\ref{fig:num_videos}. 
We see that that the tracking errors drop dramatically as we increase the number of training videos.  Our state-of-the-art results demonstrated in Section 6.1 of the main text were obtained after training on only 307 short videos, ranging from a few seconds to a few minutes in length, with an average of 52 annotations per video.  Figure~\ref{fig:num_videos} indicates that large gains could be achieved if the training set size were increased by labeling more videos.

\section{Online training}
Previous neural network trackers for tracking generic objects have been trained online~\cite{li122014deeptrack,li2015deeptrack,wang2015visual,wang2013learning,wang2015transferring,nam2015learning,zhang2015robust,danelljan2015learning,kuen2015self,hong2015tracking}.
Unfortunately, such trackers are very slow to train, ranging from 0.8 fps~\cite{li122014deeptrack} to 15 fps~\cite{wang2013learning}, with the top performing neural-network trackers running at 1 fps~\cite{nam2015learning,danelljan2015learning,kristan2015visual}.  Our tracker is trained offline in a generic manner, so no online training of our tracker is required.  As a result, our tracker is able to track novel objects at 100 fps.  

\begin{figure}[h!]
	\begin{center}
		\includegraphics[width=0.85\linewidth]{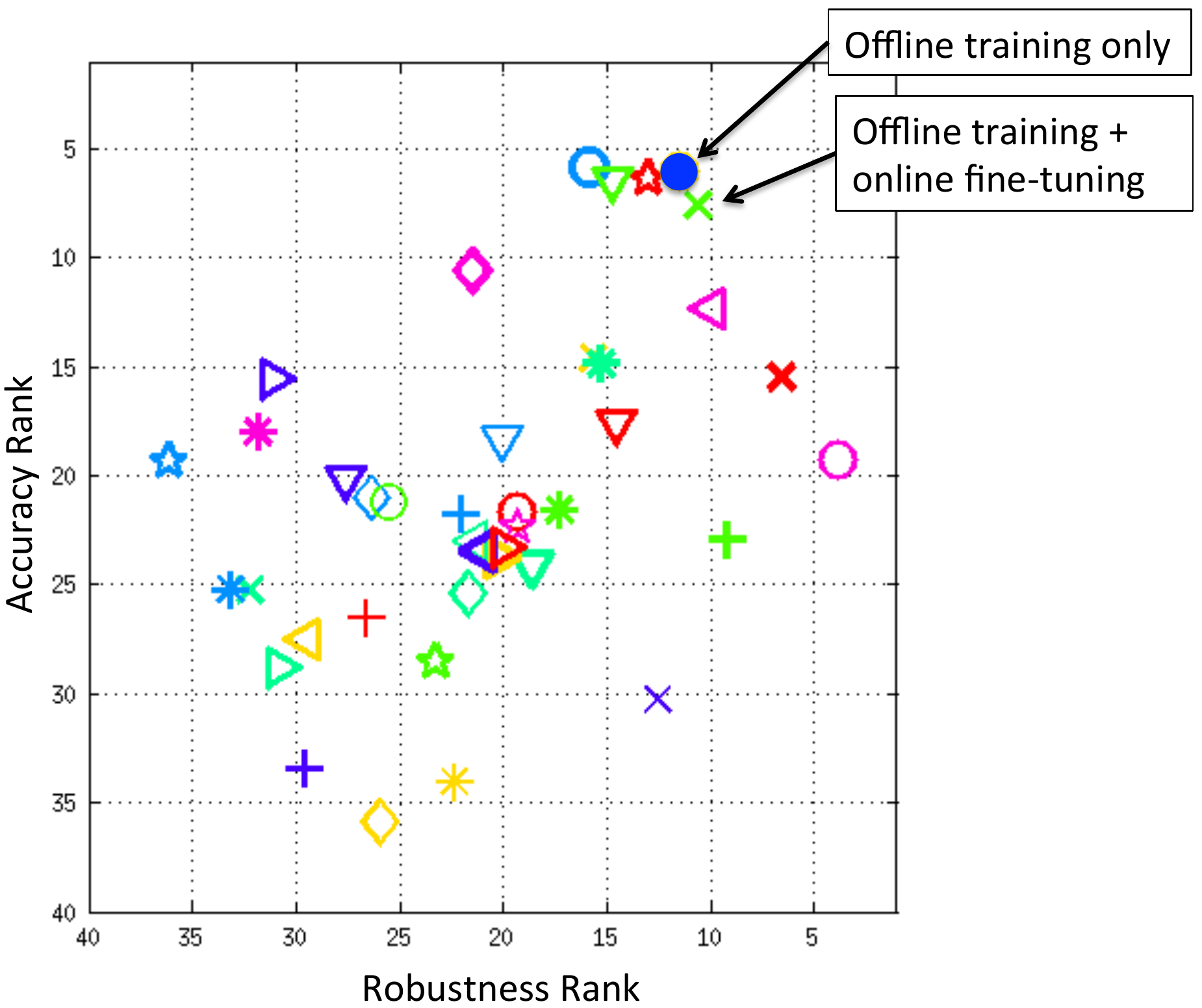}
	\end{center}
	\caption{Tracking results from the VOT 2014 tracking challenge.  Our tracker's performance is indicated with a blue circle, outperforming all previous methods on the overall rank (average of accuracy and robustness ranks).  A version of our tracker with online training is shown with a green X.  Both versions achieve approximately the same performance, demonstrating that our offline training procedure has already taught the network how to track a variety of objects.}
	\label{fig:online}
\end{figure}

\begin{figure}[h!]
	\begin{center}
		\includegraphics[width=0.95\linewidth]{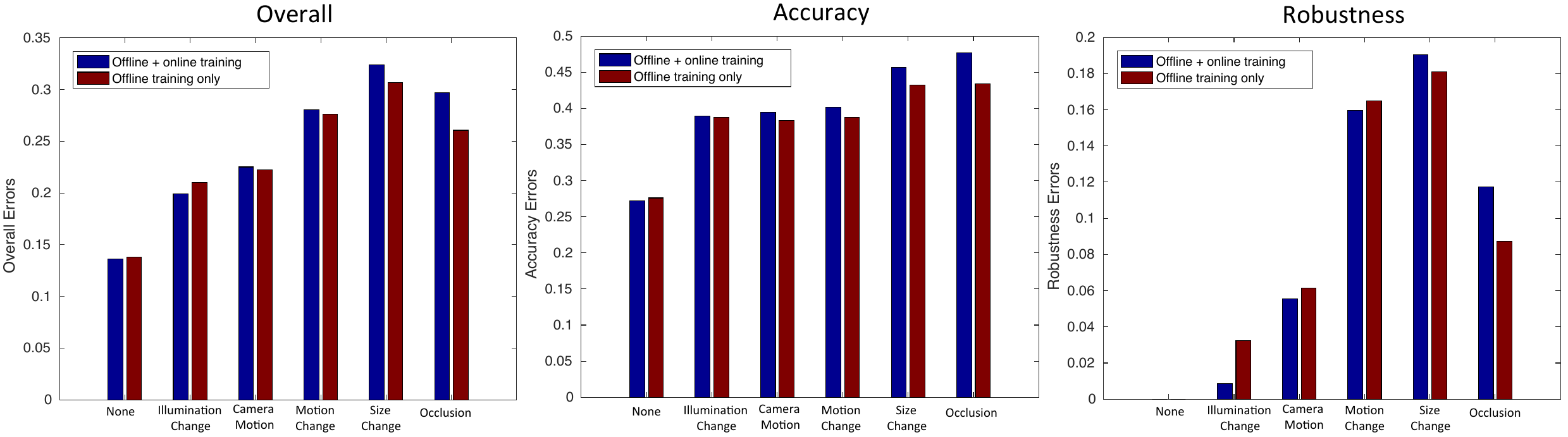}
	\end{center}
	\caption{Comparison of our tracker with and without online training (lower is better).  Both versions achieve approximately the same performance, demonstrating that our offline training procedure has already taught the network how to track a variety of objects.  Online training can lead to overfitting to the first few frames of a video, leading to more errors.}
	\label{fig:online2}
\end{figure}

In Figures~\ref{fig:online} and~\ref{fig:online2}, we explore the benefits of online training.  We use cross-validation to choose the online learning rate to be 1e-9.  Figure~\ref{fig:online} shows that online training does not significantly improve performance beyond our offline training procedure.  As might be expected, there is a small increase in robustness from online training; however, this comes at a cost of accuracy, since online training tends to overfit to the first few frames of a video and would not easily generalize to new deformations or viewpoint changes.  
A more detailed analysis is shown in Figure~\ref{fig:online2}.  

Our offline training procedure has seen many training videos with deformations, viewpoint changes, and other variations, and thus our tracker has already learned to handle such changes in a generic manner that generalizes to new objects.   Although there might be other ways to combine online and offline training, our network has already learned generic target tracking from its offline training procedure and achieves state-of-the-art tracking performance without any online training required.

\section{Generality vs Specificity}
In the main text, we analyze the generality of our tracker.  
We demonstrate that our tracker can generalize to novel objects not found in the training set.  At the same time, a user can train our tracker to track a particular class of objects especially well by giving more training examples of that class of objects.  This is useful if the tracker is intended to be used for a particular application where certain classes of objects are more prevalent.  

\begin{figure}[htb]
	\begin{center}
		\includegraphics[width=0.95\linewidth]{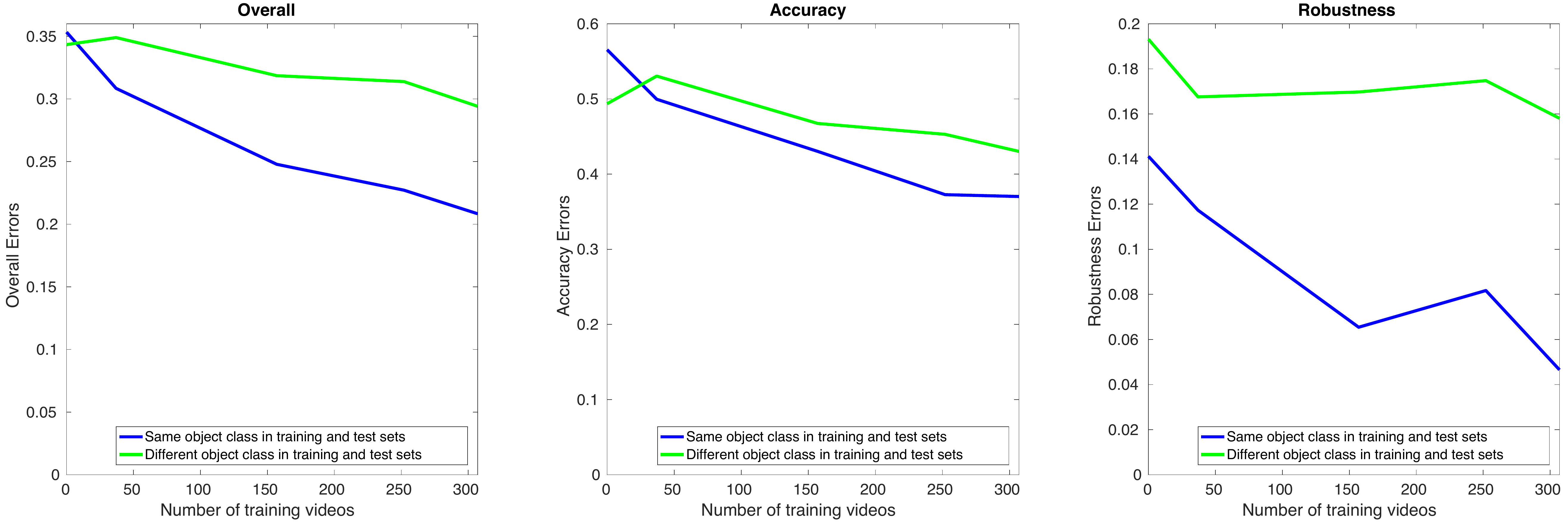}
	\end{center}
	\caption{Overall tracking errors for different types of objects in our test set as a function of the number of videos in our training set (lower is better).  Class labels are not used by our tracker; these labels were obtained only for the purpose of this analysis.}
	\label{fig:Class2}
\end{figure}

We show more detailed results of this experiment in Figure~\ref{fig:Class2}.  Analyzing the accuracy and robustness separately, we observe an interesting pattern.  As the number of training videos increases, the accuracy errors decreases equally both for object classes that appear in our training set and classes that do not appear in our training set.  On the other hand, the decrease in robustness errors is much more significant for object classes that appear in our training set compared to classes that do not.  

Thus our tracker is able to learn generic properties about objects that enable it to accurately track objects, i.e. to accurately denote the borders of the object with a bounding box.  On the other hand, the tracker's ability to generalize robustness is more limited; the tracker has a hard time tracking the motion of unknown objects when faced with difficult tracking situations.  This analysis points towards future work to increase the robustness of the tracker by labeling more videos or by learning to train on unlabeled videos.

\section{Speed analysis}
\begin{figure}[h]
	\begin{center}
		\includegraphics[width=0.95\linewidth]{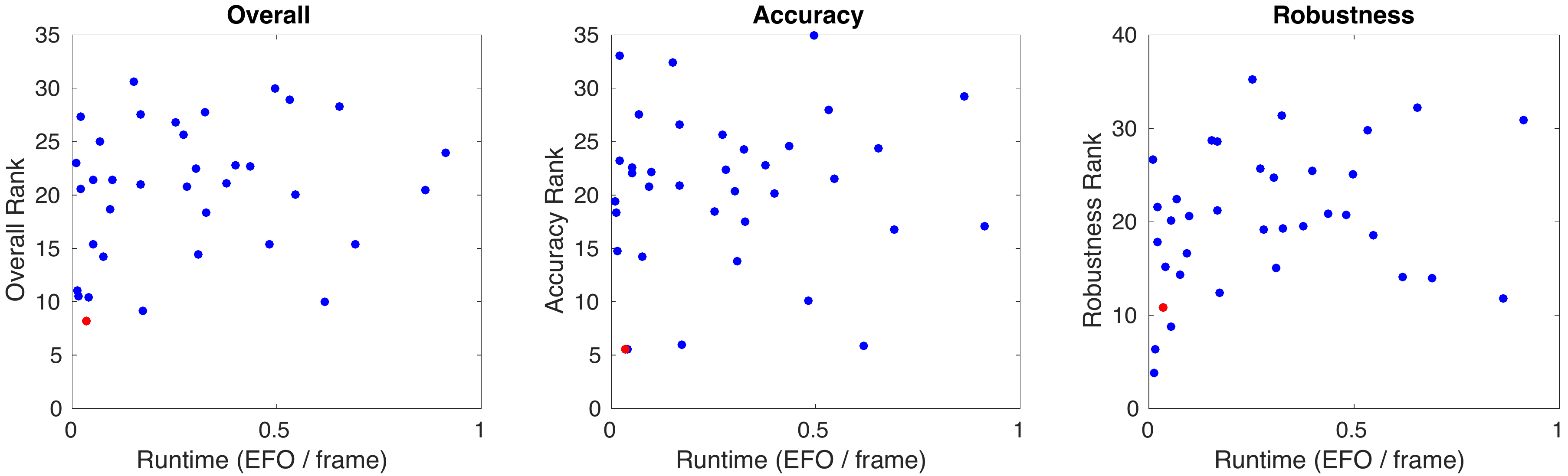}
	\end{center}
	\caption{Rank vs runtime of our tracker (red) compared to the 38 baseline methods from the VOT 2014 Tracking Challenge (blue).  Each blue dot represents the performance of a separate baseline method (best viewed in color).}
	\label{fig:speed}
\end{figure}

In the main text, we showed the speed of our tracker as a function of the overall rank (computed as the average of accuracy and robustness ranks) and showed that we have the lowest overall rank while being one of the fastest trackers.  In  Figure~\ref{fig:speed} we show more detailed results, demonstrating our tracker's speed as a function of the accuracy rank and the robustness ranks.  Our tracker has the second-highest accuracy rank, one of the top robustness ranks, and the top overall rank, while running at 100 fps.  Previous neural-network trackers range from 0.8 fps~\cite{li122014deeptrack} to 15 fps~\cite{wang2013learning}, with the top performing neural-network trackers running at only 1 fps GPU~\cite{nam2015learning,danelljan2015learning,kristan2015visual}, since online training of neural networks is slow.  Thus, by performing all of our training offline, we are able to make our neural network tracker run in real-time.

\section{How does it work?}
\begin{figure}[h]
	\begin{center}
		\includegraphics[width=0.95\linewidth]{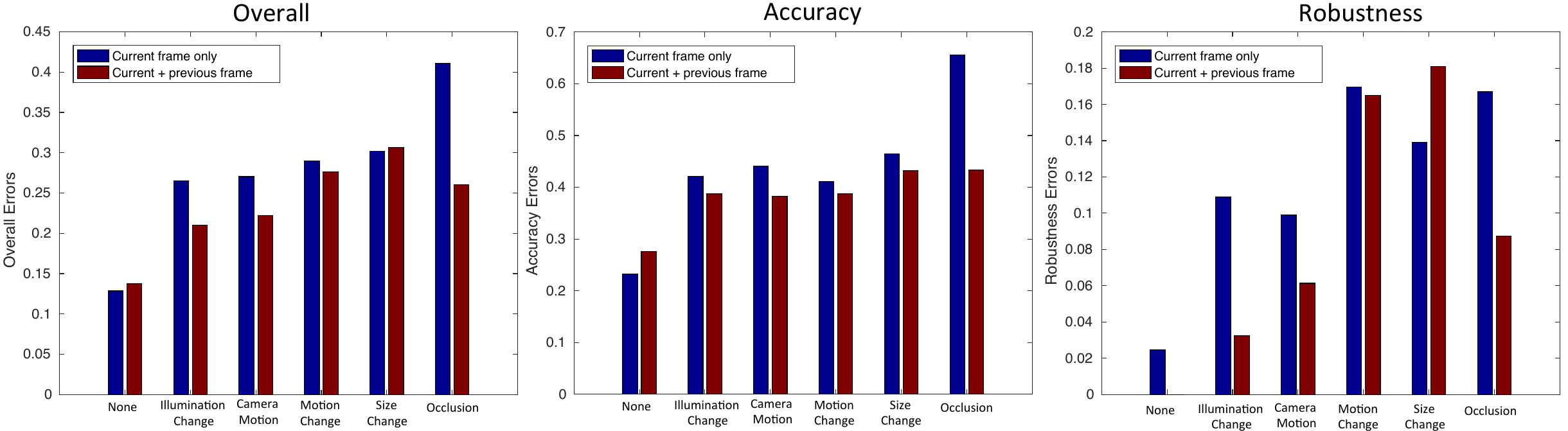}
	\end{center}
	\caption{Tracking errors for our network which receives as input both the current and previous frame, compared to a network which receives as input only the current frame (lower is better).  This comparison allows us to disambiguate between two hypotheses that can explain how our neural-network tracker works (see Section 6.2 of the main text).}
	\label{fig:memory}
\end{figure}

In the main text, we explored how our tracker works as a combination of two hypotheses:
\begin{enumerate}
	\item The network compares the previous frame to the current frame to find the target object in the current frame.
	\item The network acts as a local generic ``object detector" and simply locates the nearest ``object."
\end{enumerate}
We distinguished between these hypotheses by comparing the performance of our network to the performance of a network which does not receive the previous frame as input.  In Figure~\ref{fig:memory} we show more details of this experiment, showing also accuracy and robustness rankings.  For a more detailed interpretation of the results, see Section 6.2 of the main text.

\section{Motion Smoothness Distribution}
\label{sec:motion model track}
\begin{figure*}[bht]
	\begin{center}
		\includegraphics[width=1\linewidth]{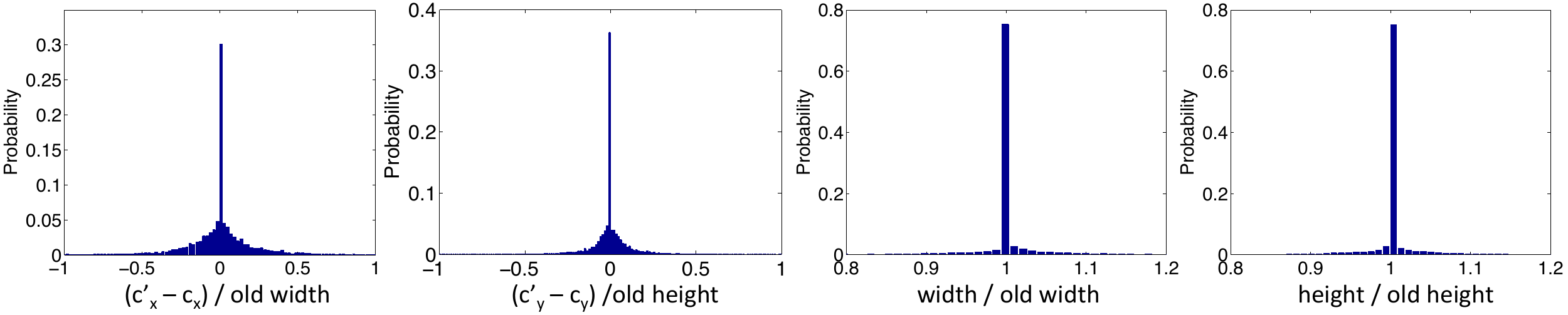}
	\end{center}
	\caption{Statistics for the change in bounding box size and location across two consecutive frames in our training set.}
	\label{fig:stats}
\end{figure*}

In Section 4.2 of the main text, we describe how we use random cropping to implicitly encode the idea that small motions are more likely than large motions.  To determine which distribution to use to encode this idea, we analyze the distribution of object motion found in the training set.  This motion distribution can be seen in Figure~\ref{fig:stats}.  As can be seen from this figure, each of these distributions can be modeled by Laplace distributions.  Accordingly, we use Laplace distributions for our random cropping procedure.  Note that the training set was only used to determine the shape of the distribution (i.e. Laplace); we use our validation set to determine the scale parameters for the distributions.

In more detail, suppose that the bounding box in frame $t-1$ is given by $(c_x, c_y, w, h)$ where $c_x$ and $c_y$ are the coordinates of the center of the bounding box and $w$ and $h$ are the width and height accordingly.  Then the bounding box at time $t$ can be seen as drawn from a distribution:
\begin{align}
c'_x &= c_x + w \cdot \Delta x \\ 
c'_y &= c_y + h \cdot \Delta y \\
w' &= w \cdot \gamma_w \\
h' &= h \cdot \gamma_h
\end{align}
with random variables $\Delta x$, $\Delta y$, $\gamma_w$, and $\gamma_h$, where $(c'_x, c'_y, w', h')$ parameterize the bounding box at time $t$ using the same representation described above.  In terms of the random variables, we can rewrite these expressions as
\begin{align}
\Delta x &= (c'_x - c_x) / w \\
\Delta y &= (c'_y - c_y) / h \\
\gamma_w &= w' / w \\
\gamma_h &= h' / h
\end{align}
The empirical distributions of these random variables over the training set are shown in Figure~\ref{fig:stats}.

\section{Number of layers}
\begin{figure}[h]
	\begin{center}
		\includegraphics[width=0.95\linewidth]{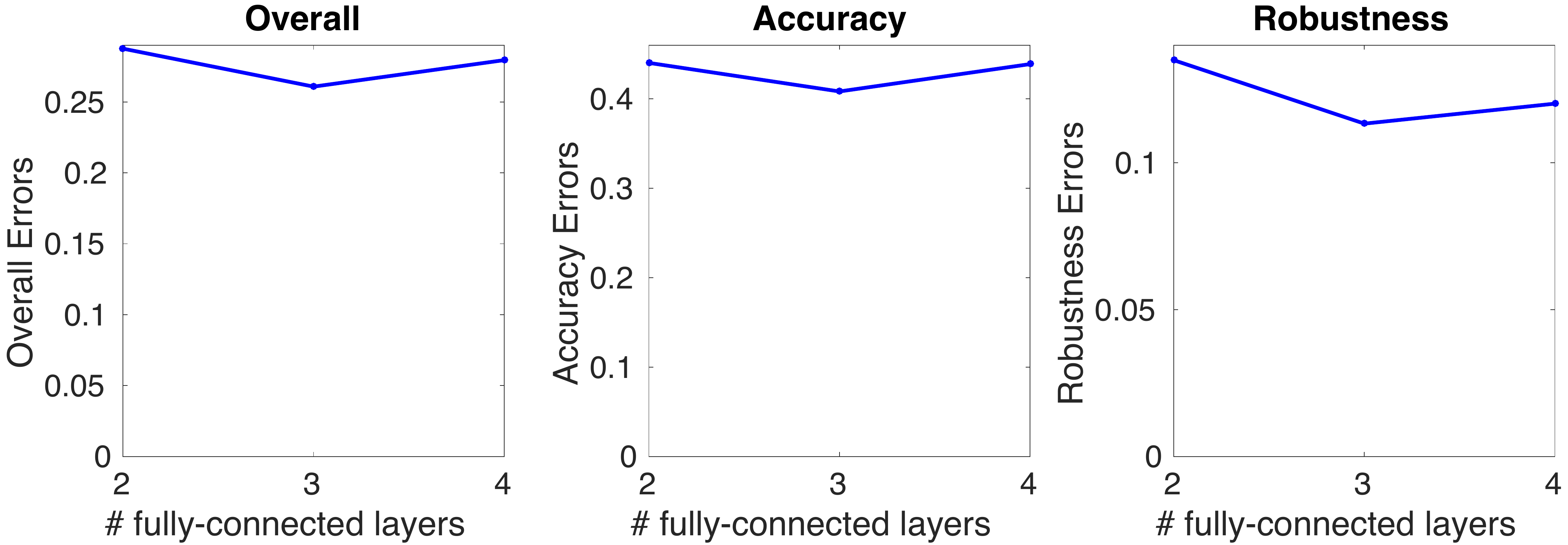}
	\end{center}
	\caption{Tracking performance as a function of the number of fully-connected layers in the neural network (lower is better).}
	\label{fig:num_layers}
\end{figure}

In Figure ~\ref{fig:num_layers} we explore the effect of varying the number of fully-connected layers on top of the neural network on the tracking performance.  These fully-connected layers are applied after the initial convolutions are performed on each image.
This figure demonstrates that using 3 fully-connected layers performs better than using either 2 or 4 layers.  However, the performance is similar for 2, 3, or 4 fully-connected layers, showing that, even though 3 fully-connected layers is optimal, the performance of the tracker is not particularly sensitive to this parameter.

\section{Data augmentation}
\begin{figure}[h]
	\begin{center}
		\includegraphics[width=0.95\linewidth]{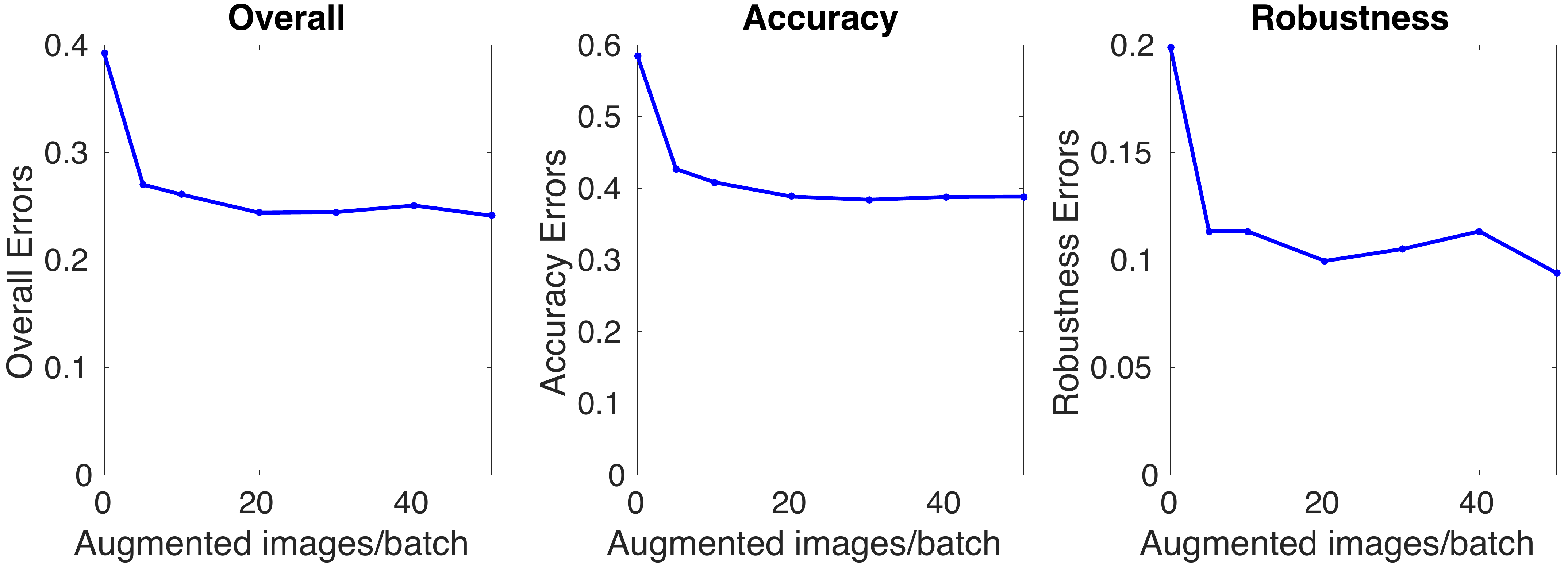}
	\end{center}
	\caption{Tracking performance as a function of the number of augmented images in each batch (lower is better).  Note that new augmented images are created on-the-fly for each batch.}
	\label{fig:amt_augmentation}
\end{figure}

In Figure~\ref{fig:amt_augmentation} we explore the effect of varying the number of augmented images created for each batch of the training set.  Note that new augmented images are created on-the-fly for each batch.  However, varying the number of augmented images varies the percentage of each batch that consists of real images compared to augmented images.  Our batch size is 50, so we can vary the number of augmented images in each batch from 0 to 49 (to leave room for at least 1 real image).

As shown in Figure~\ref{fig:amt_augmentation}, best performance is achieved when 49 augmented images are used per batch, i.e. only 1 real image is used, and the remainder are augmented.  However, performance is similar for all values of augmented images greater than 20.  In our case (with a batch size of 50), this indicates that performance is similar as long as at least 40\% of the images in the batch are augmented.  The augmented images show the same examples as the real images, but with the target object translated or with a varying scale.  Augmented images thus teach the network how the bounding box position changes due to translation or scale changes.

\section{Training Set}
\label{sec:Training_overlap}
Our training set was taken from ALOV300++~\cite{smeulders2014visual}.  
To ensure that there was no overlap with our test set, we removed 7 videos from our training set.  These videos are:
\begin{itemize}
	\item 01-Light\_video00016 
	\item 01-Light\_video00022 
	\item 01-Light\_video00023 
	\item 02-SurfaceCover\_video00012 
	\item 03-Specularity\_video00003 
	\item 03-Specularity\_video00012 
	\item 10-LowContrast\_video00013
\end{itemize}
After removing these 7 overlapping videos, there is no overlap between the videos in the training and test sets.

\section{Detailed Results}
\label{sec:rank}
The detailed results of our method compared to the 38 other methods that were submitted to the VOT 2014 Tracking Challenge~\cite{kristan2014visual} are shown in Table~\ref{tab:detailed_results}.  The VOT 2014 Tracking Challenge consists of two types of experiments.  In the first experiment, the trackers are initialized with an exact ground-truth bounding box (``exact").  In the second experiment, the trackers are initialized with a noisy bounding box, which is shifted slightly off of the target object (``noisy").  For the noisy initialization experiment, the same 15 noisy initializations are used for each tracker, and the results shown are an average of the tracking performance across these initializations.  This experiment allows us to determine the robustness of each tracker to errors in the initialization.  This noisy initialization procedure imitates that of a noisy automatic initialization process or noisy human initializations. 

The trackers are evaluated using two standard tracking metrics: accuracy and robustness~\cite{kristan2014visual,cehovin2014my}.  Each frame of the video is annotated with a number of attributes: occlusion, illumination change,  motion change, size change, and camera motion.  The trackers are ranked in accuracy and robustness separately for each attribute, and the rankings are then averaged across attributes to get a final accuracy and robustness ranking for each tracker.  The accuracy and robustness rankings are averaged to get an overall ranking, shown in Table~\ref{tab:detailed_results}.

\begin{table*}[h]
	\scriptsize
	\begin{center}
		\begin{tabular}{| p{0.2\textwidth} | c | c | c | c| c | c| c|}
			\hline
			&   \multicolumn{2}{| c |}{Overall Ranks} &  \multicolumn{2}{| c |}{Accuracy Ranks} &  \multicolumn{2}{| c |}{Robustness Ranks} & Speed \\
			\cline{2-7} 
			Method name & Exact & Noisy & Exact & Noisy & Exact & Noisy &  Frames/EFO\\
			\hline
			\textbf{GOTURN (Ours)} & \textbf{8.206944} & \textbf{8.588319} & 5.544841 & 7.227564 & 10.869048 & 9.949074 & 29.928769\\
			SAMF & 9.970153 & 9.297234 & 5.866667 & 5.685897 & 14.073638 & 12.908571 & 1.617264\\
			KCF & 10.368056 & 9.341055 & 5.533730 & 5.583333 & 15.202381 & 13.098776 & 24.226259\\
			DSST & 9.193519 & 9.393977 & 5.979630 & 5.855556 & 12.407407 & 12.932399 & 5.803051\\
			PLT\_14 & 10.526710 & 9.412576 & 14.720087 & 13.726667 & 6.333333 & 5.098485 & 62.846506\\
			DGT & 10.633462 & 9.880582 & 11.719306 & 9.318182 & 9.547619 & 10.442982 & 0.231538\\
			PLT\_13 & 11.045249 & 11.066132 & 18.340498 & 17.298932 & 3.750000 & 4.833333 & 75.915548\\
			eASMS & 14.267836 & 12.838634 & 14.220760 & 11.327036 & 14.314912 & 14.350232 & 13.080900\\
			HMMTxD & 15.398256 & 14.663101 & 10.070087 & 9.727810 & 20.726425 & 19.598391 & 2.075963\\
			MCT & 15.376874 & 15.313581 & 16.806659 & 17.576278 & 13.947090 & 13.050884 & 1.447154\\
			ABS & 19.651999 & 15.340186 & 20.666961 & 15.344515 & 18.637037 & 15.335856 & 0.623772\\
			ACAT & 14.438846 & 16.338981 & 13.796118 & 17.769841 & 15.081575 & 14.908122 & 3.237589\\
			MatFlow & 15.393888 & 16.910356 & 21.996109 & 19.142094 & 8.791667 & 14.678618 & 19.083821\\
			LGTv1 & 20.504135 & 18.189239 & 29.225131 & 26.533460 & 11.783138 & 9.845018 & 1.158273\\
			ACT & 18.676877 & 18.692439 & 20.756783 & 22.184568 & 16.596972 & 15.200311 & 10.858222\\
			VTDMG & 19.992574 & 18.835055 & 21.481942 & 20.647094 & 18.503205 & 17.023016 & 1.832097\\
			qwsEDFT & 18.365675 & 19.776101 & 17.495604 & 18.545589 & 19.235747 & 21.006612 & 3.065546\\
			BDF & 20.535189 & 19.905596 & 23.242965 & 21.731090 & 17.827413 & 18.080103 & 46.824844\\
			Struck & 21.038417 & 20.129413 & 20.868501 & 21.424688 & 21.208333 & 18.834137 & 5.953411\\
			ThunderStruck & 21.389674 & 20.333286 & 22.612468 & 21.989153 & 20.166880 & 18.677419 & 19.053603\\
			DynMS & 21.141005 & 20.479737 & 22.815739 & 21.510423 & 19.466270 & 19.449050 & 2.650560\\
			aStruck & 20.780963 & 21.465762 & 22.409722 & 20.878854 & 19.152203 & 22.052670 & 3.576635\\
			SIR\_PF & 22.705212 & 22.413896 & 24.537547 & 22.331205 & 20.872878 & 22.496587 & 2.293901\\
			Matrioska & 21.371144 & 23.119954 & 22.115980 & 21.947863 & 20.626308 & 24.292044 & 10.198580\\
			EDFT & 22.516498 & 23.176905 & 20.338931 & 22.141689 & 24.694066 & 24.212121 & 3.297059\\
			OGT & 22.463076 & 23.528818 & 14.810633 & 16.893364 & 30.115520 & 30.164271 & 0.393198\\
			CMT & 22.788164 & 23.852773 & 20.098007 & 22.612765 & 25.478321 & 25.092781 & 2.507500\\
			FoT & 23.003472 & 24.375915 & 19.388889 & 21.623392 & 26.618056 & 27.128439 & 114.643138\\
			IIVTv2 & 25.669987 & 24.610138 & 25.651061 & 25.400309 & 25.688913 & 23.819967 & 3.673112\\
			IPRT & 25.014620 & 25.081882 & 27.564283 & 26.643535 & 22.464957 & 23.520229 & 14.688296\\
			PTp & 27.288300 & 25.208133 & 33.046296 & 30.268937 & 21.530303 & 20.147328 & 49.892214\\
			LT\_FLO & 23.958402 & 26.020573 & 17.075617 & 20.843334 & 30.841186 & 31.197811 & 1.096522\\
			FSDT & 28.275770 & 26.805519 & 24.378835 & 24.318730 & 32.172705 & 29.292308 & 1.529770\\
			IVT & 28.892955 & 27.820781 & 27.952576 & 27.432765 & 29.833333 & 28.208796 & 1.879526\\
			IMPNCC & 27.566645 & 29.293698 & 26.570580 & 29.349962 & 28.562711 & 29.237434 & 5.983489\\
			CT & 30.585835 & 29.377864 & 32.462103 & 30.823647 & 28.709566 & 27.932082 & 6.584306\\
			FRT & 27.800316 & 29.554293 & 24.300128 & 27.199856 & 31.300505 & 31.908730 & 3.093665\\
			NCC & 26.831924 & 30.305656 & 18.497180 & 23.444646 & 35.166667 & 37.166667 & 3.947948\\
			MIL & 30.007762 & 30.638921 & 34.934175 & 35.527778 & 25.081349 & 25.750064 & 2.012286\\
			\hline
		\end{tabular}
	\end{center}
	\caption{Full results from the VOT 2014 tracking challenge, comparing our method (GOTURN) to the 38 other methods submitted to the competition.  We initialize the trackers in two different ways: with the exact ground-truth bounding box (``Exact") and with a noisy bounding box (``Noisy").}
	\label{tab:detailed_results}
\end{table*}

\end{document}